\documentclass[journal,comsoc]{IEEEtran}
\usepackage[T1]{fontenc}
\ifCLASSINFOpdf
\else

\fi
\usepackage{float}
\usepackage{hyperref}
\usepackage{amsmath}
\usepackage{color}
\usepackage[dvipsnames]{xcolor}
\usepackage[export]{adjustbox}
\usepackage{color}
\usepackage{graphicx}
\usepackage{booktabs}
\usepackage{multirow}
\usepackage{amsmath}
\usepackage{array}
\usepackage{graphicx}
\usepackage{cite}
\usepackage{xcolor}
\definecolor{darkred}{RGB}{237,21,21}
\definecolor{darkgreen}{RGB}{21,189,21}
\begin{document}

\title{GRAD-Former: Gated Robust Attention-based Differential Transformer for Change Detection}

\author{Durgesh Ameta\textsuperscript{*}, Ujjwal Mishra\textsuperscript{*}, Praful Hambarde and Amit Shukla,~\IEEEmembership{Member,~IEEE}

\thanks{\textsuperscript{*}These authors contributed equally to this work.\\
Durgesh Ameta is with Indian Knowledge System and Mental Health Applications (IKSMHA) Center, IIT Mandi, Mandi, HP 175005, India (durgesgameta@gmail.com).\\
Ujjwal Mishra is with School of Computing, IIIT Una HP-177209, India (ujjwalmishra238@gmail.com).\\
Praful Hambarde and Amit Shukla are with Center of Artificial Intelligence and Robotics (CAIR), IIT Mandi, Mandi, HP 175005, India (praful@iitmandi.ac.in, amitshukla@iitmandi.ac.in).\\}
}
\maketitle

\begin{abstract}
Change detection (CD) in remote sensing aims to identify semantic differences between satellite images captured at different times. While deep learning has significantly advanced this field, existing approaches based on convolutional neural networks (CNNs), transformers and Selective State Space Models (SSMs) still struggle to precisely delineate change regions. In particular, traditional transformer-based methods suffer from quadratic computational complexity when applied to very high-resolution (VHR) satellite images and often perform poorly with limited training data, leading to under-utilization of the rich spatial information available in VHR imagery.
We present GRAD-Former, a novel framework that enhances contextual understanding while maintaining efficiency through reduced model size. The proposed framework consists of a novel encoder with Adaptive Feature Relevance and Refinement (AFRAR) module, fusion and decoder blocks. AFRAR integrates global-local contextual awareness through two proposed components: the Selective Embedding Amplification (SEA) module and the Global-Local Feature Refinement (GLFR) module. SEA and GLFR leverage gating mechanisms and differential attention, respectively, which generates multiple softmax heaps to capture important features while minimizing the captured irreverent features.
Multiple experiments across three challenging CD datasets  (LEVIR-CD, CDD, DSIFN-CD) demonstrate GRAD-Former's superior performance compared to existing approaches. Notably, GRAD-Former outperforms the current state-of-the-art models across all the metrics and all the datasets while using fewer parameters. Our framework establishes a new benchmark for remote sensing change detection performance. Our code will be released at: \href{https://github.com/Ujjwal238/GRAD-Former}{\textcolor{blue}{https://github.com/Ujjwal238/GRAD-Former}}
\end{abstract}

\begin{IEEEkeywords}
 
Change detection (CD), high-resolution remote
sensing image, differential attention, transformer.

\end{IEEEkeywords}
\IEEEpeerreviewmaketitle
\section{Introduction}
\label{sec:I}
\IEEEPARstart{T}{he} Earth's surface undergoes continuous transformation due to both natural processes and human activities, necessitating a reliable and systematic monitoring methodology to assess these changes \cite{1}. Change detection (CD) is the use of remote sensing multi-temporal data to monitor a geographic area and to identify and describe the change features of phenomena of interest or objects, to quantitatively analyze this difference. CD serves as an essential tool in fields such as urban land management \cite{2}, \cite{3}, sustainable resource utilization \cite{4}, and emergency response decision-making \cite{5}, providing more efficient alternatives to traditional manual field investigations.

\begin{figure}[t]
\begin{center}

   \includegraphics[width=1\linewidth]{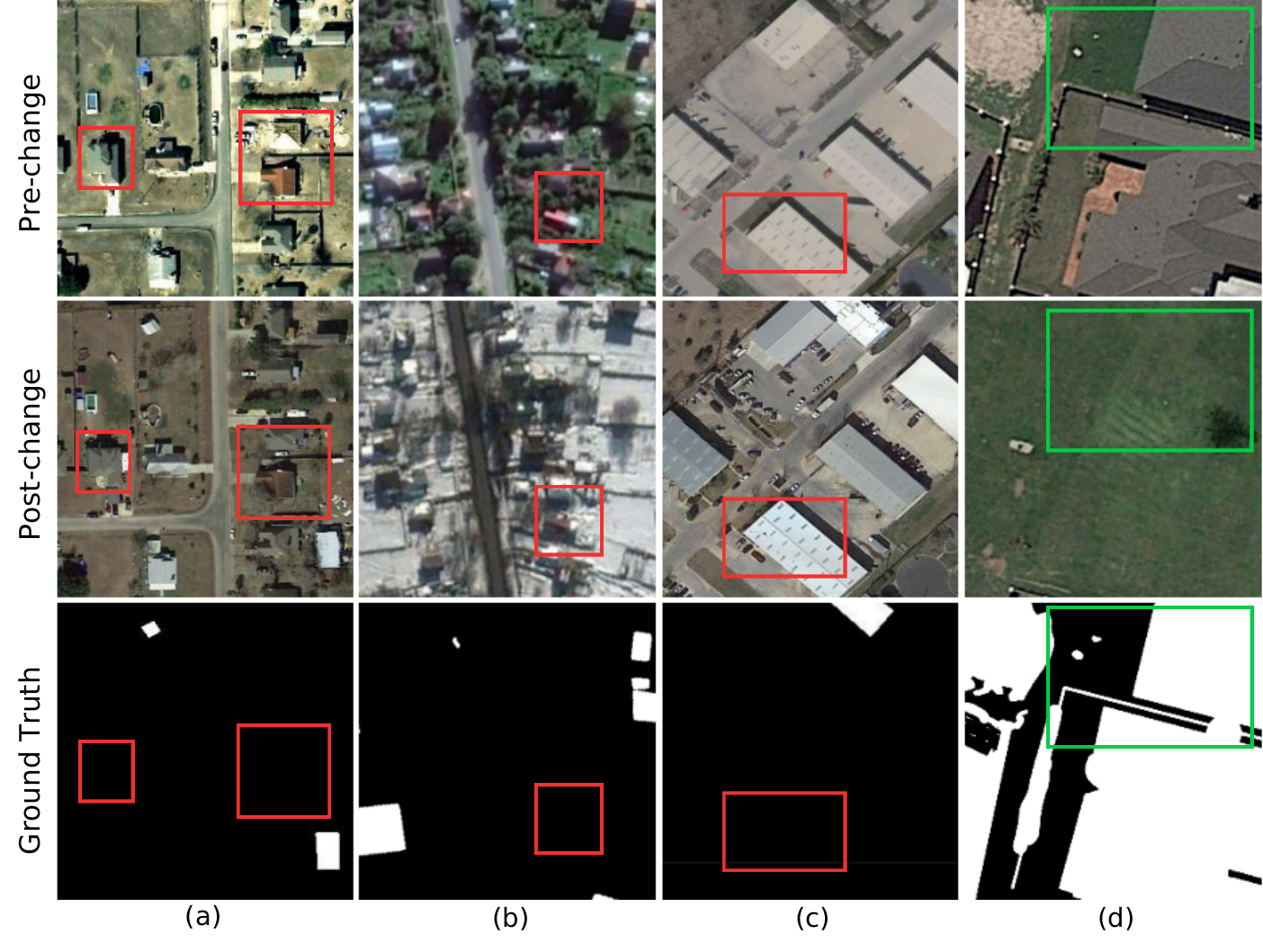} %
\end{center}
   \caption{Example images showing different challenges in the CD task, along with their ground-truth change masks. It is especially difficult to ignore irrelevant changes (marked in red boxes), such as (a) shadows and lighting differences, (b) seasonal variations, and (c) moving cars and roof changes, while accurately detecting (d) both large and small meaningful changes (marked in the green box).}
   \vspace{-2pt}
\label{fig:Fig1}
\end{figure}

Advancements in satellite technology have improved both the quality and diversity of remote sensing data. However, the performance gains of existing models and techniques have not kept pace with these enhancements. As resolution increases, noise and background information become more evident \cite{6}. Moreover, bitemporal data suffer from illumination disturbances and seasonal variations between images, as depicted in Fig. \ref{fig:Fig1}(a) and Fig. \ref{fig:Fig1}(b). In addition, shifts in movable objects such as vehicles, alterations in roof colors, and significant variations in the size of changed areas are often observed, with multiple adjacent change regions leading to highly inconsistent features, as visible in Fig. \ref{fig:Fig1}(c) and Fig. \ref{fig:Fig1}(d), which makes it very challenging to distinguish between the relevant and the non-relevant features.

Thus, to accurately identify the change regions, researchers have developed many traditional representative CD algorithms, such as discriminant projection based methods \cite{new}, slow feature analysis \cite{7}, change vector analysis \cite{8} and transformation based techniques \cite{9}. However, these classical approaches failed to capture the diverse features and neglected spatial contexts \cite{10}. This attributed to the development of deep learning based model that uses convolutional autoencoders (CAEs) \cite{11}, pixel-based deep neural networks (DNNs) \cite{12}, recurrent neural networks (RNNs) \cite{13}, convolutional neural networks (CNNs) \cite{14} and demonstrated robustness to data noise and grasped spatio-temporal contextual information efficiently. 

As deep learning methods remain dominant, Siamese architectures continue to be widely used to address the CD problem \cite{DLreview}. Typically, these methods incorporate architectural elements like multi-scale feature extraction, dilated convolutions and attention mechanisms into their framework. CNN-based architectures \cite{15}, \cite{17}, \cite{18} have long been the standard approach for the CD task. However, they are recently being replaced by transformer-based architectures \cite{16}, \cite{19}, \cite{20} that use attention mechanisms \cite{21} to address CNN's difficulty in capturing global contextual representations. Although simple transformer-based architectures offer advantages over CNNs, they also have notable limitations. They encounter difficulties in capturing local contextual information, which is essential for accurately segmenting subtle variations along complex and irregular boundaries. Additionally, their computational complexity scales quadratically, leading to a high number of parameters and substantial memory requirements. This makes them less suitable for practical remote sensing change detection applications. Therefore, an optimal approach should aim to eliminate unnecessary features while capturing important local and global contextual information to detect both significant and structural changes between image pairs with minimal parameters.

To address the aforementioned challenges, we propose GRAD-Former, highly efficient change detection framework. The key novelty of our design is the Adaptive Feature Relevance And Refinement (AFRAR) module, which effectively filters out noise and irrelevant background information in Very High Resolution (VHR) satellite images, allowing the model to focus solely on the essential local and global contextual details. Our approach maximizes the use of high-resolution optical imagery by capturing both fine-grained local features and broad global patterns with high precision.  

To achieve this, AFRAR module is equipped with Selective Embedding Amplification module (SEA), which employs gated techniques for robust feature extraction, ensuring that only relevant information is retained. Additionally, our Global-Local Feature Refinement (GLFR) module leverages differential attention \cite{differentialtransformer} to extract global contextual information while capturing minimal noise and reducing computational overhead.
We rigorously evaluate GRAD-Former on three diverse change detection benchmark datasets, demonstrating its superiority over the State-Of-The-Art (SOTA) methods. Our model achieves the best-in-class performance, as evidenced by its outstanding quantitative and qualitative results on publicly available datasets.  

The main contributions of our study are summarized as follows:  
\begin{itemize}
    \item We propose GRAD-Former, a robust Siamese-based CD framework that efficiently mitigates noise and irrelevant background information to accurately detect semantic differences between bitemporal satellite images.
    \item We introduce the concept of differential attention combined with gated mechanisms for change detection in the GLFR module and SEA module in the AFRAR module, along with multi-scale difference-based fusion in Differential Amalgamation (DA) module, which integrates difference features with encoded features to enhance focus on change regions.
    \item Extensive experiments on three public change detection datasets validate the effectiveness of our approach, demonstrating its robustness and efficiency, achieving state-of-the-art performance.
    
\end{itemize}

The rest of this article is organized as follows. Section ~\ref{sec:II} provides an overview of related work. Section ~\ref{sec:III} details the GRAD-Former architecture proposed for change detection. The experimental analysis of the model is given in Section ~\ref{sec:IV}. Finally, Section ~\ref{sec:V} concludes this article.

\section{Related Work}
\label{sec:II}
In this section, we review various approaches for the change detection task, including CNN-based, transformer-based, and State Space Models (SSMs)-based methods, along with their respective advantages and limitations.

\subsection{CNN-Based CD methods}
Among the various deep learning architectures, Convolutional Neural Networks (CNNs) have emerged as a powerful tool due to their ability to effectively extract spatial and hierarchical features from images. CNNs excel at capturing local information through convolutional filters, making them particularly well-suited for pixel-wise tasks such as semantic segmentation and remote sensing change detection (CD). This capability has led to their widespread adoption in numerous fields in computer vision, including medical image segmentation \cite{medi}, object detection \cite{obj_detec}, and environmental monitoring \cite{env_moni}.

Early approaches leveraged Fully Convolutional Networks (FCNs) and U-Net architectures \cite{unet}, which provided a strong foundation by employing an encoder-decoder structure to extract multi-scale spatial features. However, U-Net suffered from feature loss due to its linear compression. To address this, enhancements such as skip connections, as seen in UNet++ \cite{unetpp}, were used to refine feature fusion and retention. SNUnet \cite{snu}. BiDateNet \cite{bidate}, based on the U-Net framework, demonstrated improvements in change detection by integrating multi-scale feature representations. 

The evolution of CNN-based change detection models saw the emergence of Siamese networks, which effectively processes bitemporal remote sensing images by employing dual-stream architectures. Daudt \textit{et al.} \cite{daudt} introduced three Siamese-based models: FC-EF, FC-Siam-Conc, and FC-Siam-Di, each implementing different fusion strategies to compare temporal changes effectively. Additionally, Khan \textit{et al.} \cite{khan} proposed a weakly supervised CNN-based approach to directly localize change detection without requiring densely labeled data. 

Despite their advantages, CNN-based methods exhibit notable limitations. They struggle with capturing long-range dependencies \cite{cnn_back}, limiting their ability to model temporal and spatial relationships effectively. Moreover, their sensitivity to varying illumination, cloud cover, and environmental conditions poses challenges in real-world change detection tasks.

\subsection{Transformer-Based CD methods}
Transformers have significantly advanced computer vision (CV), excelling in various tasks such as image classification, segmentation, and object detection. Their powerful ability to model global contextual relationships has made them particularly effective for change detection (CD) tasks in remote sensing (RS) \cite{swin}. Unlike CNNs, which primarily capture local spatial structures, transformers aggregate both local and global features through self-attention mechanisms, improving feature representation across bitemporal images \cite{bit}.

Early transformer-based CD models evolved from UNet-like architectures, leveraging hierarchical feature extraction. For instance, SwinSUNet \cite{swin} integrates a hierarchical Swin Transformer structure within a Siamese UNet framework to enhance local and global feature fusion. But this approach uses a very large number of parameters. To improve multi-scale feature utilization, methods like TFI-GR \cite{tfir} and A2Net \cite{a2net} employ lightweight transformer-based architectures with supervised attention mechanisms to highlight change areas. Meanwhile, BiTemporal attention transformers \cite{bitemporal} leverage cross-attention mechanisms to refine the interaction of bitemporal features for more precise change localization. 

Other methodologies have combined the strengths of CNNs and transformers. Hybrid networks, such as Hybrid Transformer \cite{hybrid} and MFiNet \cite{mfi}, fuse CNN's local feature extraction with transformer's global contextual modeling, achieving superior CD performance, Noman \textit{et al.} in ELGC-Net \cite{elgcnet} achieves this by integrating CNN and PT-attention in a single module, each working on half the channels. ChangeFormer \cite{19} enhances long-term dependency modeling, ensuring robust feature extraction across multiple temporal scales. Additionally, methods like Trans-MAD \cite{transmad}
and WSMsFNet \cite{wsm} integrate transformer-based spatial and channel attention mechanisms to refine the perception of change areas. Wu \textit{et al.}'s recent work, CGMNet \cite{gated_hyper} explores gated mechanism based attention networks for CD task in hyperspectral images. Recent research in CD has further explored the development of plug-and-play frameworks. In particular, AdvCP \cite{adver} proposes an adversarial class prompting scheme designed to uncover co-occurring pixel-level representations learned by image-level classification models under weakly supervised learning settings with SAM (Segment Anything Model) based baseline.

Despite of transformer's ability to capture long term dependencies, they struggle to find the relevant features in VHR bitemporal images due to excessive noise and high information redundancy present in them.

\subsection{SSM-Based CD methods}
State Space Models (SSMs) have evolved from control system theory, where they were initially developed to model input-response relationships in linear time-invariant systems. Their introduction to computer vision and remote sensing stems from the need to efficiently process long-range sequence data while mitigating the computational complexity of traditional models. Unlike Transformers, which exhibit quadratic complexity due to self-attention, SSMs, such as Mamba, offer a linear time complexity, making them highly scalable for large-scale visual tasks \cite{mamba}.

Mamba and its variants have been successfully applied in a range of tasks, including image classification, object detection, and medical image segmentation. Vision Mamba (VMamba) extends Mamba’s capabilities by incorporating a four-directional scanning mechanism (top-left, bottom-right, top-right, bottom-left), allowing efficient traversal of spatial features while preserving computational efficiency \cite{vmamba}. Similarly, RS-Mamba introduces a multi-directional scanning approach to enhance feature extraction in remote sensing images \cite{rs}. 

In the field of change detection (CD), Mamba-based approaches such as CDMamba \cite{cdmam} and LCCDMamba \cite{LCD} have been developed to better capture spatial-temporal differences between bitemporal images. CDMamba employs residual structures and local-global feature extraction to detect changes effectively, while LCCDMamba further integrates multi-scale fusion techniques to preserve high-dimensional semantics. ChangeMamba builds on VMamba’s scanning mechanisms to enhance spatial-temporal feature relationships, demonstrating significant advancements in dense prediction tasks \cite{changemamba}. Recent advancements include SpectralMamba \cite{spectral} and S2Mamba \cite{s2}, which integrate spectral-spatial fusion techniques for hyperspectral change detection, as well as CM-UNet \cite{cmu} and RS3Mamba \cite{rs3}, which enhance feature alignment across temporal frames for improved accuracy. Additionally, emerging works such as LocalMamba \cite{localmam} focus on dynamically optimizing scanning schemes for various layers, addressing the trade-off between efficiency and fine-grained feature extraction. Qu \textit{et al.} in their recent work, MambaFedCD \cite{mamba_cd} explores federated active learning strategy with SSM backbone for CD task.

Despite these advancements, SSMs have notable limitations. While they excel in capturing long-range dependencies, they often struggle with fine-grained local feature extraction, which is crucial for accurate change detection. Most existing methods focus on modifying scanning techniques or leveraging pre-trained backbones rather than addressing the inherent challenges of local representation. Additionally, SSM-based models may suffer from sensitivity issues in tasks requiring precise boundary delineation, limiting their applicability in high-resolution remote sensing.

\section{Methodology}

\label{sec:III}

 Our base framework is a transformer-based Siamese network. It consists of three main components: an encoder, fusion module, and a decoder \cite{bit}. The Siamese encoder processes pre- and post-change images, extracting multi-scale feature maps across four stages. The extracted feature maps at each stage, denoted as $\hat{\mathcal{F}}_{pre}^i$ and $\hat{\mathcal{F}}_{post}^i$, where $i \in [1,2,3,4] $, are then passed to the fusion module to achieve bitemporal feature interaction. The fused features from all stages are fed into the decoder, where they are combined and refined through multiple convolutional and transposed convolutional layers to generate high-resolution feature maps. These upsampled feature maps are then processed by the prediction layer to produce the final change map.
\par
Further, this section describes the overall architecture of the network and explains the AFRAR module in detail.
\subsection{Overall Architecture}

\begin{figure*}[t!]
\begin{center}
   \includegraphics[width=1\linewidth]{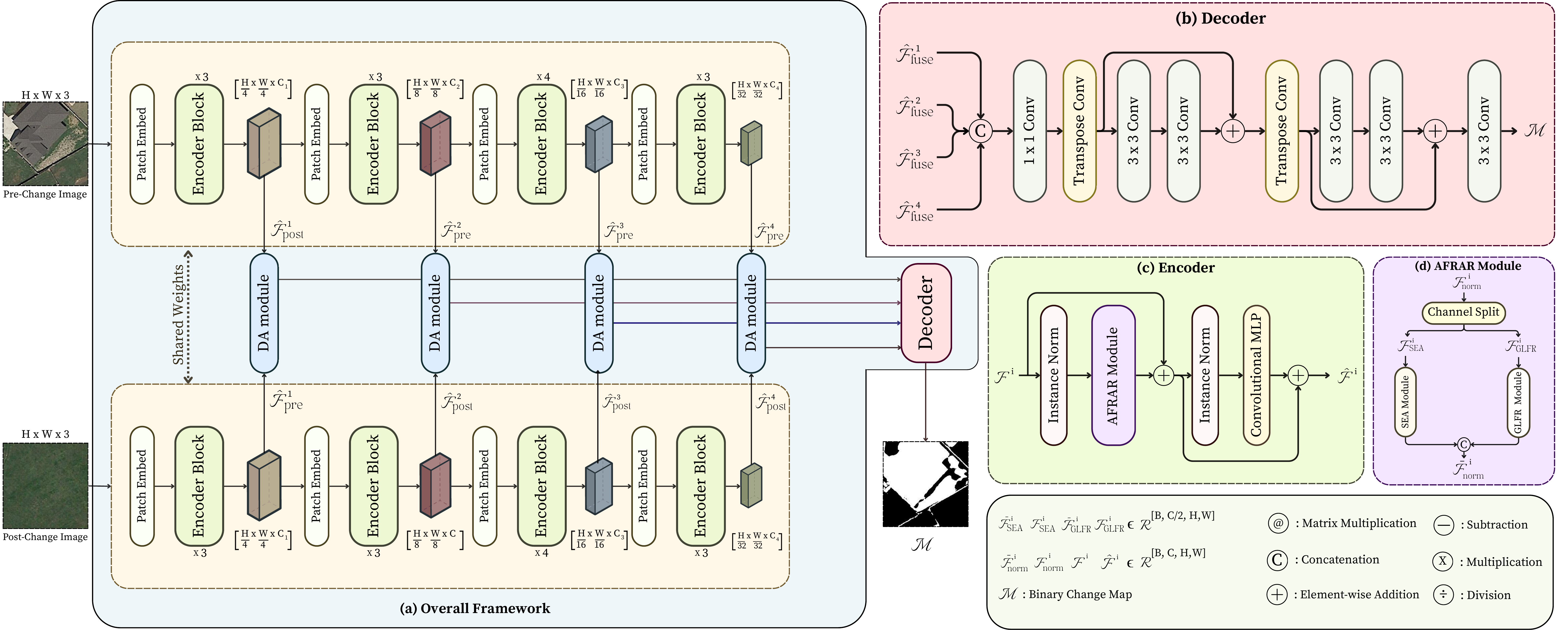}
\end{center}
\vspace{-4mm}
   \caption{The overall design of the proposed GRAD-Former is depicted as follows: (a) The full network structure is outlined, showcasing the pre- and post-change input images processed through a shared encoder. This encoder extracts features across four stages, which are then combined using DA module. At each stage, the fused feature maps are merged within the decoder, which employs multiple convolutional and transposed convolution layers to generate upsampled features. These upsampled feature maps are ultimately utilized in the prediction layer to produce the final change map. (b) Detailed architecture of the decoder, it processes the fused feature maps \( \hat{F}_{fuse}^{i} \) extracted from the four stages (\( i = 1, 2, 3, 4 \)) by concatenating them along the channel dimension. It then applies a \( 1 \times 1 \) convolution to project the features. Subsequently, upsampling is performed twice to restore the spatial dimensions to match the model input, utilizing a cascaded transpose convolution followed by a residual block consisting of two convolutional layers. Finally, a convolutional layer is employed to generate the prediction scores.
      (c) illustrates the encoder block structure, incorporating the novel Adaptive Feature Relevance and Refinement (AFRAR) module alongside a convolutional MLP layer. Instance normalization is applied across channels before input is fed into the AFRAR and MLP layers.
   (d) The detailed architecture of the AFRAR module is shown, which first splits the input ${\mathcal{F}}_{norm}^i$ along the channel and combines SEA and GLFR modules to extract relevant features, whose outputs are concatenated to form $\bar{\mathcal{F}}_{norm}^i$.
 }
\label{fig:overallarch}
\end{figure*}

\begin{figure}[h]
\begin{center}
   \includegraphics[width=0.75\linewidth]{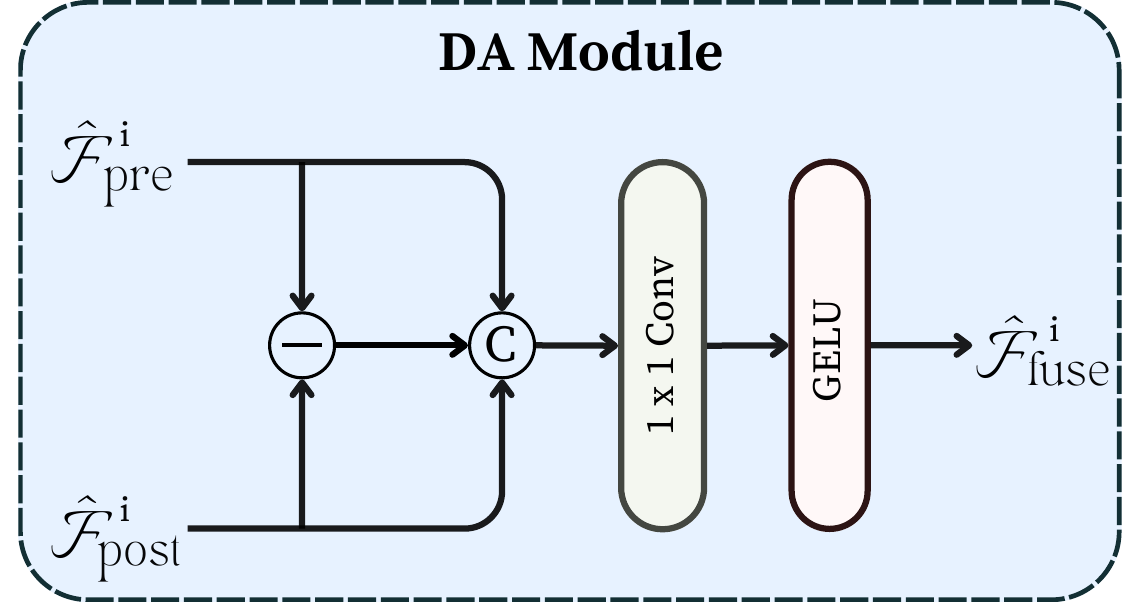}
\end{center}
   \caption{Detailed illustration of the proposed Difference Amalgamation (DA) module that captures semantic and differential features by concatenating \( \hat{\mathcal{F}}_{pre}^i \) and \( \hat{\mathcal{F}}_{post}^i \) along the channel dimension, along with their difference. Next, a convolutional layer is applied to reduce the channel count, followed by an activation layer within the DA module, producing \( \hat{\mathcal{F}}_{fuse}^i \), where \( i \in [1,2,3,4] \).)).
}
   \vspace{-6px}
\label{fig:DA}
\end{figure}

The architecture of the proposed change detection (CD) framework, referred to as GRAD-Former, is illustrated in Fig. \ref{fig:overallarch}(a). Similar to the base framework, GRAD-Former consists of a encoder, fusion module, and a decoder. The encoder takes a pre and post-change satellite images as input and generates multi-scale feature maps $\hat{\mathcal{F}}_{pre}^i$ and $\hat{\mathcal{F}}_{post}^i$, where $i \in [1,2,3,4] $. At each stage, the feature maps are first down-sampled using a patch embedding layer and then processed through a series of encoder blocks. The structure of the encoder block is depicted in Fig. \ref{fig:overallarch}(c), it includes the proposed novel Adaptive Feature Relevance And Refinement (AFRAR) module along with a convolutional MLP layer, instance normalization across the channel is performed before passing the input to the AFRAR module and convolution MLP layer. Fig. \ref{fig:overallarch}(d) depicts the proposed AFRAR module in detail, containing the SEA and the GLFR modules for relevant feature extraction and concatenating their respective outputs to get $\bar{\mathcal{F}}_{norm}^i$.

\begin{figure*}[t]
\begin{center}
   \includegraphics[width=1\linewidth]{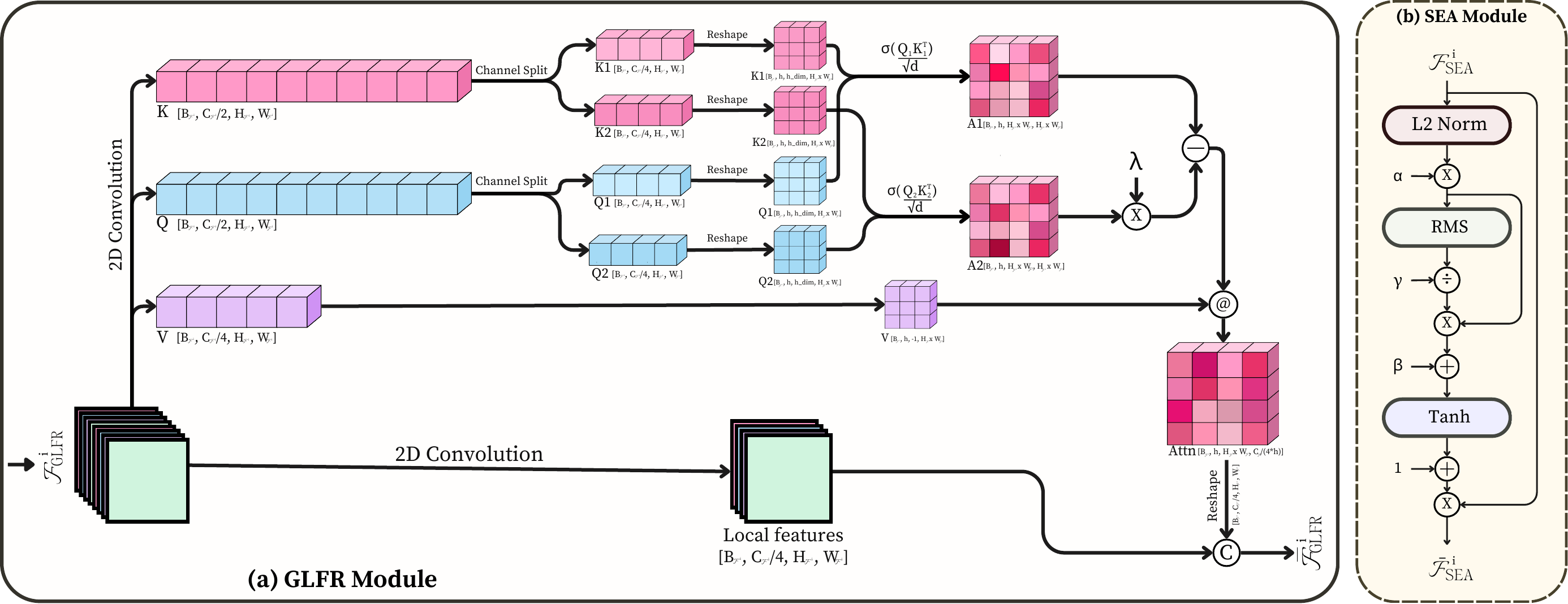}
\end{center}
\vspace{-4mm}
   \caption{(a) Overview of the proposed Global-Local Feature Refinement (GLFR) Module, An attention mechanism designed to overcome the diffused focus problem in transformers by implementing differential multi-head attention. The module generates $Q$, $K$, $V$ matrices from input $\mathcal{F}^i_{\text{GLFR}}$ and splits $Q$, $K$ into pairs to calculate dual softmax attention maps ($A_1$, $A_2$). By computing the difference $A=A_1-\lambda \cdot A_2$ with learnable scaling factor $\lambda$, the module creates sparse attention patterns that focus exclusively on relevant features while filtering out noise. The resulting attention output is combined with local features, providing an efficient balance between global context and local detail without the computational burden typical of transformer architectures.
   (b) depicts the proposed Selective Embedding Amplification (SEA) module that selectively amplifies important features through gating mechanism. The input features $\mathcal{F}^i_{\text{SEA}}$ undergo $L2$ normalization and multiplication by learnable parameter $\alpha$. Using another learnable parameter $\gamma$, a normalization factor is computed. The gating function ($G(x)=1+\tanh(x+\beta)$) then adaptively weights each channel based on its importance where $\beta$ is another learnable scalar. Then the final output $\bar{\mathcal{F}}^i_{\text{SEA}}=\mathcal{F}^i_{\text{SEA}} \cdot G$ enhances relevant features while suppressing noise, making the model robust against the sparsity of information in high-resolution satellite imagery. 
}
\label{fig:novelmodules}
\end{figure*}
 
At each stage, in line with the baseline framework, feature pairs extracted from the encoder blocks undergo fusion through the proposed novel \textbf{Differential Amalgamation} (DA) modules. As described in Fig. \ref{fig:DA}, the DA module encodes the semantic and differential features by concatenating $\hat{\mathcal{F}}_{pre}^i$ and $\hat{\mathcal{F}}_{post}^i$ across the channel dimension along with their difference. Subsequently, a convolutional layer is employed to reduce the number of channels, followed by an activation layer in the DA module giving $\hat{\mathcal{F}}_{fuse}^i$ where $i \in [1,2,3,4] $ as an output. This can be summarized as:
\begin{equation}
\hat{\mathcal{F}}_{\text{fuse}}^i = \sigma\left( \text{Conv}_{1 \times 1}\left( \text{Conc}\left( \hat{\mathcal{F}}_{\text{pre}}^i,\ \hat{\mathcal{F}}_{\text{post}}^i,\ [\hat{\mathcal{F}}_{\text{post}}^i - \hat{\mathcal{F}}_{\text{pre}}^i] \right) \right) \right)
\end{equation}

where, $ Conc\ {(Concatination)}$ is performed along the channel dimension, and $\sigma(\cdot)$ denotes the GELU activation function.

This $\hat{\mathcal{F}}_{fuse}^i$ is then fed to to a decoder to predict the binary change map ($\mathcal{M}$).

\par \textbf{Decoder Architecture:} GRAD-Former employs a decoder consisting of multiple convolution and transpose convolution layers, as shown in Fig. \ref{fig:overallarch}(b). The multi-scale fused features ($\hat{\mathcal{F}}_{fuse}^i$ where $i \in [1,2,3,4] $) from four stages are concatenated along the channel dimension and passed through a $1 \times 1$ convolution layer. Next, transpose convolution is applied to progressively increase the spatial resolution of the feature maps. To further refine these features, a residual block containing two $3 \times 3$ convolution layers is incorporated. This combination of transpose convolution and residual blocks is applied twice to restore the spatial dimensions to match the input size. Finally, a convolution layer generates prediction scores with two output channels: one representing the no-change class and the other corresponding to the change class. The final binary change map is obtained by applying the argmax operation along the channel dimension.
\par
Next, we present a detailed discussion of our Adaptive Feature Relevance And Refinement (AFRAR) module.
\subsection{{Adaptive Feature Relevance And Refinement (AFRAR) module}}


The AFRAR module is designed to effectively capture relevant, global and local contextual information and ignore the noise and irrelevent background information while minimizing computational complexity compared to traditional self-attention. Drawing inspiration from the inception module \cite{inception}, which utilizes parallel convolution operations with different kernel sizes, our approach divides the input channels into separate groups, applying distinct operations to each split (see Fig. \ref{fig:overallarch}(d)). Specifically, the input features are partitioned channel-wise and processed through two independent context aggregation branches to extract important, global and local contextual representations. In the heart of our AFRAR module there are two novel Selective Embedding Amplification (SEA) (Fig. \ref{fig:novelmodules}(b)) and Global-Local Feature Refinement (GLFR) (Fig. \ref{fig:novelmodules}(a)) modules that specialize in filtering and capturing essential information along with the global-local context.
\par
Let $\mathcal{F}^i \in \mathcal{R}^{H^i \times W^i \times C^i}$  
be the input feature to the encoder block. After normalization it becomes $\mathcal{F}^i_{norm} $ which serves as the input to the AFRAR at the $i^{th}$ stage,  
where $(H^i, W^i, C^i)$ denotes the input image's height, width, and channels, respectively.  

We first perform a channel-wise split on the normalized input feature $\mathcal{F}^i_{\text{norm}}$,  
resulting in $\mathcal{F}^i_{\text{GLFR}}$ and $\mathcal{F}^i_{\text{SEA}}$, where  
$\mathcal{F}^i_{\text{GLFR}}, \mathcal{F}^i_{\text{SEA}} \in \mathcal{R}^{H^i \times W^i \times \frac{C^i}{2}}$.  
These serve as inputs to our SEA and GLFR module, respectively. That is,  
\begin{equation}
\mathcal{F}^i_{\text{GLFR}}, \mathcal{F}^i_{\text{SEA}} = \text{Split}(\mathcal{F}^i_{norm}).
\end{equation}
\par
Next, we describe the SEA and GLFR modules within our AFRAR module in detail.
\subsubsection{\textbf{Selective Embedding Amplification (SEA) module}} 
Inspired from \cite{gated1}, \cite{gated2}, \cite{gated3}, \cite{gated4} that use gated mechanisms in their decoder module to enhance the quality of the input mapping to the output, but these models lack the involvement of the key features in the input itself. Aiming to enhance the expressive capability of channel features while minimizing the number of model parameters, we introduce SEA module in our encoder block to maximize the quality of captured information. The input features to the SEA, $\mathcal{F}^i_{\text{SEA}}$ are first subject to $L2$ normalization and are then multiplied by a learnable parameter $ \alpha $. This parameter is used to learn and evaluate importance of each channel. During the L2-norm
processing , a constant $\epsilon$ ($0\leq \epsilon \leq 10^{-5}$) is
introduced to prevent numerical instability. This embedding ($E$) is given by:
\begin{equation}
E = \alpha \cdot \sqrt{\sum_{H,W} {\mathcal{F}^i_{\text{SEA}}}^2 + \epsilon}
\end{equation}
After computing embeddings $E$, we find the $RMS$ value of $E$ and introduce another learnable parameter, $\gamma$ that learns the competitive relationships between channels. We divide this $\gamma$ with the calculated RMS to obtain a normalization factor $N$.
\begin{equation}
N = \frac{\gamma}{\sqrt{\mathbb{E}_c [E^2] + \epsilon}}
\end{equation}
where \( \mathbb{E}_c [\cdot] \) represents the mean across the channel dimension. $\epsilon$ is introduced for numeric stability. Then another learnable parameter $\beta$ is introduced to  amplify
the contribution of important channels and reduce the impact of irrelevant ones. This $\beta$ is used in the gating function as,
\begin{equation}
G = 1 + \tanh(E \cdot N + \beta)
\end{equation}
where $G$ is the output of the gating function. Choosing $1+tanh(x)$ as the gating function gives a nonlinear adjustment of the channel weights, which avoids gradient explosion. Finally we get the output of the SEA $\bar{\mathcal{F}}_{SEA}^i$ by multiplying $G$ with the input $\mathcal{F}^i_{\text{SEA}}$ element-wise.
\begin{equation}
\bar{\mathcal{F}}_{SEA}^i = {\mathcal{F}}_{SEA}^i \cdot G
\end{equation}
This $G$ matrix contains all the important and relevant features and are thus amplified in the output adding robustness to our model. It becomes necessary to encode these features as VHR images could carry large amounts of noise and relevant information could be very sparse.

\subsubsection{\textbf{Global-Local Feature Refinement (GLFR) module}}
The other half of the $\mathcal{F}^i_{\text{norm}}$ are processed by the module as $\mathcal{F}^i_{\text{GLFR}}$ $\in \mathcal{R}^{H^i \times W^i \times \frac{C^i}{2}}$. Key issue with existing transformer based CD methods is that they tend to spread their attention too thin, often focusing on irrelevant information, which impacts performance, leading to slower processing, irrelevant outputs, this only increases as the quality of satellite image increases due to the added additional background information and noise. To solve this issue, we propose GLFR module.  Inspired by \cite{differentialtransformer}, we use multi-head differential attention to capture the global and relevant information effectively. To achieve this we first get the Query(Q), Key(K) and the Value(V) matrices by:
\begin{equation}
Q = W_Q \mathcal{F}^i_{\text{GLFR}}, \quad K = W_K \mathcal{F}^i_{\text{GLFR}}, \quad V = W_V \mathcal{F}^i_{\text{GLFR}}
\end{equation}
where \( Q, K \in \mathcal{R}^{B \times C/2 \times H \times W} \) and \( V \in \mathcal{R}^{B \times C/4 \times H \times W} \). In a traditional transformer, the attention mechanism distributes focus across all tokens in a sequence using a single softmax function. This often results in important content being drowned out by irrelevant context, especially in long or complex inputs. To counter this we calculate two softmax maps, one to focus on relevant tokens and the other to represent the noise or distractions in the input and their difference would result in sparse attention patterns, where the attention scores become concentrated only on the relevant tokens, making the model more efficient and accurate. We do this by splitting the $Q$ and $K$ matrices along the channel into $Q_1, Q_2$ and $K_1,K_2$ respectively and then calculating the respective attention scores as:
\begin{equation}
Q_1, Q_2 = \text{split}(Q), \quad K_1, K_2 = \text{split}(K)
\end{equation}
here \( Q_1,Q_2,K_1,K2 \in \mathcal{R}^{B \times C/4 \times H \times W} \) which, are then reshaped for attention to \( \mathcal{R}^{B \times h \times h_{dim} \times (H\times W)} \) where $h$ is the number of attention heads that is chosen to be 4 in our model and $h_{dim}$ represents the dimension (number of channels) per attention head, which is adjusted as per the input as:
\begin{equation}
    h_{dim}= \left\lfloor \frac{input\ dimension}{4 \times head}\right\rfloor
\end{equation}
The Value(V) matrix is also reshaped such that $V \in \mathcal{R}^{B \times h \times -1 \times (H\times W)} $. We calculate softmax heaps $A_1$ and $A_2$ for $Q_1,K_1 $ and $ Q_2,K2$ respectively as:
\begin{equation}
A_1 = \sigma\left(\frac{Q_1^T K_1}{\sqrt{h_{dim}}}\right), \quad A_2 = \sigma\left(\frac{Q_2^T K_2}{\sqrt{h_{dim}}}\right)
\end{equation}
\text{where } $\sigma(x)$ \text{ is the softmax function.} Now, the difference between the obtained softmax heaps should contain the relevant information while suppressing overlapping noise. This subtraction is analogous to how noise-canceling headphones remove ambient noise by canceling out common sound waves. We get this final softmax heap ($A$) by:
\begin{equation}
\text{A} = \left( A_1 - \lambda \cdot (A_2) \right) 
\end{equation}
where $A \in \mathcal{R}^{B \times h \times (H\times W) \times (H\times W)}$. $\lambda$ is a learnable scalar that controls the influence of the second softmax map (representing noise). It is calculated as:
\begin{equation}
\lambda = \lambda_1 - \lambda_2 + \lambda_{\text{init}}
\end{equation}
where,
\begin{equation}
\lambda_1 = e^{(\lambda_{q_1} \cdot \lambda_{k_1}})
\end{equation}
\begin{equation}
\lambda_2 = e^{(\lambda_{q_2} \cdot \lambda_{k_2}})
\end{equation}

where \( \lambda_{\text{init}} = 0.8 \) and $\lambda_{q_1}, \ \lambda_{k_1},  \ \lambda_{q_2}, \ \lambda_{k_2} \in \mathcal{R}^{h_{dim}}$ \text{are learnable vectors}. Finally, we find the $Attn$ matrix:
\begin{equation}
Attn = \text{A} @  V^T
\end{equation}
where $@$ is the matrix multiplication operation.
$Attn$ is then reshaped to $ \mathcal{R}^{B \times C/4 \times H \times W} $ and concatenated with local features ($l_{feat}$) obtained by
\begin{equation}
    l_{feat} = reshape(\mathcal{F}^i_{\text{GLFR}})
\end{equation}
$l_{feat} \in \mathcal{R}^{B \times C/4 \times H \times W} $.
We overcome the computational overhead problem of transformers by only calculating the attention on a very reduced channel size (see Fig. \ref{fig:novelmodules}(a)) and doing it effectively giving more robust and error-free results. The output $\bar{\mathcal{F}}^i_{\text{GLFR}}$ of this module is given by concatenating $l_{feat}$ and $Attn$ across the channel.
\begin{equation}
\bar{\mathcal{F}}_{GLFR}^i = \text{concat}(l_{\text{feat}}, Attn)
\end{equation}

\begin{table*}[t]
\centering
\renewcommand{\arraystretch}{1.2}

\caption{Performance comparison of different methods on CDD, DSIFN-CD, and LEVIR-CD datasets. Best results are in \textcolor{darkgreen}{green} and second-best results are in \textcolor{darkred}{red}.}
\begin{tabular}{|c|c|c|ccc|ccc|ccc|}
\hline
\multirow{2}{*}{\textbf{Type}} & \multirow{2}{*}{\textbf{Method}} & \multirow{2}{*}{\textbf{Publication}} & \multicolumn{3}{c|}{\textbf{CDD}} & \multicolumn{3}{c|}{\textbf{DSIFN-CD}} & \multicolumn{3}{c|}{\textbf{LEVIR-CD}} \\
\cline{4-12}
& & & \textbf{$F_1$} & \textbf{IoU} & \textbf{OA(\%)} & \textbf{$F_1$} & \textbf{IoU} & \textbf{OA(\%)} & \textbf{$F_1$} & \textbf{IoU} & \textbf{OA(\%)} \\
\hline
\multirow{7}{*}{CNN-based} & FC-EF \cite{18} & \textit{ICIP-2018} & 65.38 & 48.57 & 93.07 & 64.38 & 47.47 & 87.55 & 81.16 & 68.29 & 98.12 \\
& FC-Sima-diff \cite{18} & \textit{ICIP-2018} & 70.37 & 54.29 & 94.31 & 62.03 & 44.96 & 88.17 & 83.77 & 72.08 & 98.44 \\
& FC-Sima-conc \cite{18} & \textit{ICIP-2018} & 72.65 & 57.05 & 94.47 & 65.42 & 48.61 & 87.60 & 86.26 & 75.85 & 98.60 \\
& DTCDSCN \cite{DTCDSCN} & \textit{LGRS-2021} & 94.14 & 88.93 & 98.63 & 87.21 & 76.58 & 95.59 & 89.75 & 81.41 & 98.98 \\
& SNUNet \cite{snu} & \textit{LGRS-2022} & 95.34 & 91.11 & {98.90} & 79.64 & 66.17 & 93.48 & 88.59 & 79.51 & 98.85 \\
& ISNet \cite{isnet}& \textit{TGRS-2022} & 95.63 & 91.84 & 98.35 & 78.42 & 67.01 & 88.69 & 90.32 & 82.35 & \textcolor{darkred}{99.04} \\
& SARAS-Net \cite{saras} & \textit{AAAI-2023} & 93.19 & 87.33 & 98.26 & 67.58 & 51.04 & 89.01 & 88.37 & 79.16 & 98.84 \\
\hline
\multirow{3}{*}{Mamba-based} & ChangeMamba \cite{changemamba} & \textit{TGRS-2024} & 94.91 & 90.32 & {98.81} & \textcolor{darkred}{90.21} & \textcolor{darkred}{82.17} & 94.39 & 90.18 & 82.07 & 99.01 \\
& RS-Mamba \cite{rs} & \textit{LGRS-2024} & 93.59 & 87.96 & 98.49 & 86.67 & 76.48 & 95.56 & 89.56 & 81.06 & 98.96 \\
& CDMamba \cite{cdmam} & \textit{arXiv-2024} & 94.84 & 90.20 & 98.83 & 85.81 & 75.13 & 95.17 & 90.41 & 82.25 & 99.02 \\
& MF-VMamba \cite{mf-vmamba} & \textit{TGRS-2024} & 95.69 & 91.58 & 98.97 & 88.05 & 78.73 & \textcolor{darkred}{96.01}& 90.64 & 82.89 & 99.07 \\
\hline
\multirow{5}{*}{Transformer-based} & BIT \cite{bit}& \textit{TGRS-2022} & 94.37 & 89.34 & 98.68 & 71.49 & 55.64 & 90.74 & 89.23 & 80.55 & 98.94 \\
& ChangeFormer \cite{chnageformer} & \textit{IGARSS-2022} & 94.21 & 89.06 & 98.63 & 87.34 & 77.53 & {95.76} & {90.46} & {82.58} & \textcolor{darkred}{99.04} \\
& ScratchFormer \cite{scratchformer} & \textit{TGRS-2024} & \textcolor{darkred}{96.12} & \textcolor{darkred}{92.54} & \textcolor{darkred}{99.09} & 67.62 & 45.13 & 89.74 & 88.94 & 80.08 & 98.90 \\
& ELGC-Net \cite{elgcnet} & \textit{TGRS-2024} & 92.32 & 85.74 & 98.22 & 73.76 & 58.42 & 92.68 & 89.11 & 80.36 & 98.88 \\
& Meta-SGNet \cite{meta-sg} & \textit{TGRS-2024} & - & - & - & - & - & - & 89.20 & 80.50 & 98.93 \\  
& RAHFF-Net \cite{rahff} & \textit{JSTARS-2025} & 94.46 & 87.96  & 98.64 & - & - & - & 89.41 & 79.72 & 98.92 \\
& AMDANet \cite{amda} & \textit{TGRS-2025} & - & - & - & - & - & - & 90.32 & 82.34 & \textcolor{darkred}{99.04} \\  
& CICD \cite{cicd} & \textit{JSTARS-2025} & - & - & - & - & - & - & \textcolor{darkred}{91.23} & \textcolor{darkred}{83.87} & - \\  
& AdvCP \cite{adver} & \textit{TIP-2026} & - & - & - & 58.90 & 43.10 & 85.75 & 70.06 & 50.45 & 98.95 \\  
& \textbf{GRAD-Former} & \textit{-} & \textcolor{darkgreen}{97.57} & \textcolor{darkgreen}{95.26} & \textcolor{darkgreen}{99.43} & \textcolor{darkgreen}{93.14} & \textcolor{darkgreen}{87.16} & \textcolor{darkgreen}{97.65} & \textcolor{darkgreen}{91.52} & \textcolor{darkgreen}{84.36} & \textcolor{darkgreen}{99.14} \\
\hline
\end{tabular}
\label{tab:ALL}
\end{table*}

\section{Experimental Analysis}
\label{sec:IV}
This section provides a comprehensive overview of the datasets, evaluation metrics and experimental setup. It also analyzes GRAD-Former's performance across multiple datasets, compares it to previous state-of-the-art models through both qualitative and quantitative evaluations, and presents ablation studies to assess the importance of each component in our model.
\subsection{Datasets}
In this study, three publicly available datasets with uneven sample distribution were selected, namely, the LEVIR-CD dataset \cite{16}, the DSIFN-CD dataset \cite{17}, and the CDD dataset \cite{cdd} in order to completely validate the efficacy of GRAD-Former.
\subsubsection{LEVIR-CD \cite{16}}
This dataset is a large-scale benchmark for building change detection, comprising 637 high-resolution image pairs with a spatial resolution of 0.5 meters per pixel and a size of 1024 × 1024 pixels. These images, sourced from Google Earth, capture changes over a time span ranging from 5 to 14 years. The dataset specifically focuses on detecting building construction and demolition. For our experiments, we utilize a cropped version of the dataset, where non-overlapping image patches of 256 × 256 pixels are extracted. Following the standard protocol outlined in \cite{chnageformer}, we adopt the default data split, consisting of 7120 image pairs for training, 1024 for validation, and 2048 for testing.
\subsubsection{DSIFN-CD \cite{dsifnd}}
This binary change detection dataset, sourced from Google Earth, comprises high-resolution bitemporal images from six major cities in China: Xi’an, Chongqing, Beijing, Chengdu, Wuhan, and Shenzhen. It includes a total of 3988 remote sensing image pairs, capturing various land cover changes such as water bodies, cultivated land, buildings, and roads. The dataset is available in cropped sub-image pairs of 256 × 256 pixels and is split into training, validation, and test sets with 14,400, 1,360, and 192 image pairs, respectively.
\subsubsection{CDD \cite{cdd}}
The CDD dataset is a publicly available collection obtained from Google Earth, consisting of 11 raw image pairs with resolutions ranging from 0.03 to 1 m/pixel. Among these, seven pairs depict seasonal changes (4,725 × 2,700), while four pairs capture disaster-related transformations (1,900 × 1,000), covering diverse geographic areas and landscapes. Each pair includes complex variations such as cloud cover, shadows, lighting differences, and rotations. The dataset primarily focuses on seasonal shifts in natural elements, from individual trees to extensive forested regions, with a smaller portion highlighting structural changes, from vehicles to large buildings. It is also available in a cropped 256 × 256 format, comprising 10,000, 3,000, and 3,000 image pairs for training, validation, and testing, respectively.

\begin{figure*}[t]
\begin{center}
   \includegraphics[width=1\linewidth]{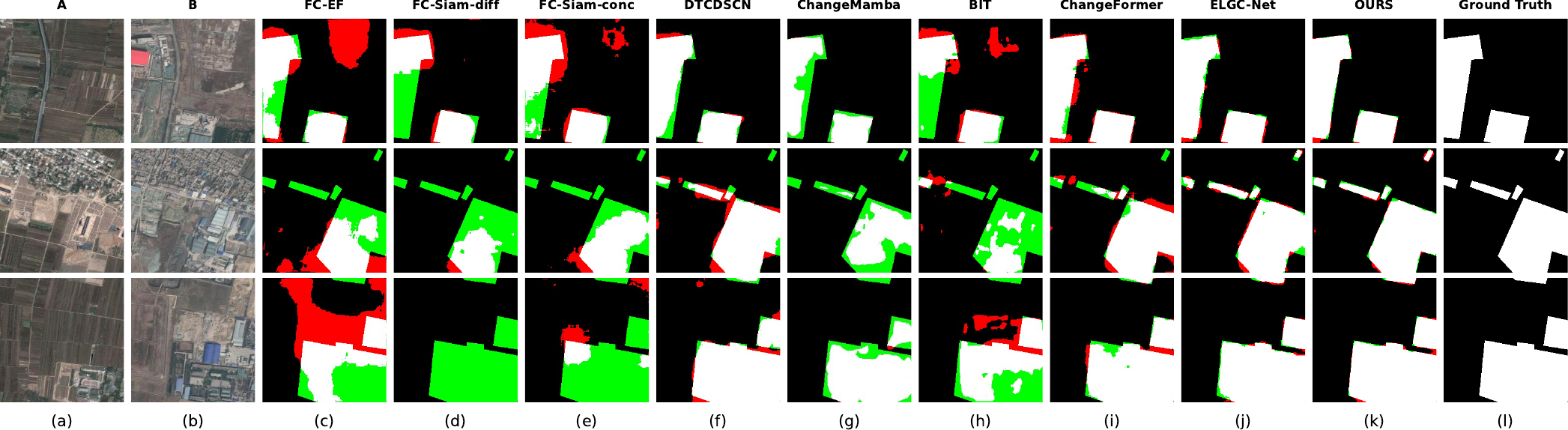}
\end{center}
   \caption{Qualitative comparisons on the DSIFN-CD dataset. We show the comparison with the eight best existing change detection
approaches in the literature, whose codebases are publicly available.(a) Pre-Change image (A), (b) Post-Change image (B), (c) FC-EF \cite{18}, (d) FC-Siam-diff \cite{18}, (e) FC-Siam-conc \cite{18}, (f) DTCDSCN \cite{DTCDSCN}, (g) ChangeMamba \cite{changemamba}, (h) BIT \cite{bit}, (i) ChangeFormer \cite{bit}, (j) ELGC-Net \cite{elgcnet}, (k) GRAD-Former (OURS) and (l) Ground truth.  Color scheme: white
represents TP (i.e., “changed”), black corresponds to TN (i.e., “unchanged”), \textcolor{darkred}{red} indicates FP, and \textcolor{darkgreen}{green} denotes FN. }
\label{fig:dsifn}
\end{figure*}

\subsection{Evaluation Metrics}
We evaluate the change detection results using intersection over union (IoU), $F_1$ score, and overall accuracy (OA(\%)) on all datasets, where IoU and $F_1$ score are calculated for the change class.
\subsection{Experimental Setup}
We have implemented the proposed GRAD-Former using PyTorch\footnote{https://pytorch.org} deep learning framework and powered by a single NVIDIA
RTX A4000 GPU with 16 GB memory during the training and testing phase. The AdamW optimizer is utilized for model optimization, with an initial learning rate set to 0.0001. The learning rate follows a linear decay policy, with scheduled decay at 100 and 200 epochs. The training process is conducted over 400 epochs, ensuring comprehensive learning of the model. The batch size is configured at 2 to accommodate training constraints. For model evaluation, validation is performed at each epoch, and the best model is selected based on validation performance to ensure optimal generalization. The testing phase follows using a dedicated evaluator, assessing model performance on the test set.

The GRAD-Former encoder consists of four stages, structured with 3, 3, 4, and 3 encoder blocks, respectively, generating feature maps with channel dimensions of 64, 96, 128, and 256. The encoder processes input image pairs of size 256 × 256 × 3. On the other hand, the decoder produces a binary change map that maintains the original spatial resolution. To enhance model generalization, data augmentation techniques—including random flipping, scaling, cropping, color jittering, and Gaussian blur—are applied during training. During inference, the model weights remain frozen, and the input image pair is passed through the network. The model generates a probability change map with two channels, from which a binary change map is derived using the argmax operation along the channel dimension.

\subsubsection{Loss Function}During the training phase, we employ a pixel-wise cross-entropy loss to assess the network's performance. The cross-entropy loss function is mathematically defined as:
\begin{equation}
\text{Loss} = -\frac{1}{N} \sum_{i=1}^{N} \left[ y_i \log(\hat{y}_i) + (1 - y_i) \log(1 - \hat{y}_i) \right]
\end{equation}
where \( y \) represents the actual class label, \( \hat{y} \) denotes the predicted probability, and \( N \) refers to the total number of pixels.

\subsection{Quantitative Comparison with State-of-The-Art Methods}
We present a quantitative comparison of our approach with
state-of-the-art (SOTA) methods with open code bases over LEVIR-CD, DSIFN-CD,
and CDD-CD datasets on key metrics ($F_1$,
OA(\%), and IoU) across last 8 years in Table \ref{tab:ALL}. However, due to the unavailability of official codebases of the latest models (RAHFF-Net \cite{rahff}, AMDANet \cite{amda}, Meta-SGNet \cite{meta-sg}, CICD \cite{cicd}, DgFA \cite{dgfa}) at the time of our study, we were unable to conduct a full evaluation across all datasets and metrics. Instead, we report the results as presented in their respective papers to ensure a fair and representative comparison on all the datasets that were common with our study. To highlight the methods with greater
detection performance, the best, second best values in each evaluation metric are given in \textcolor{darkgreen}{green} and \textcolor{darkred}{red}. Table \ref{tab:method_comparison} compares our model's parameter count, GFLOPS and $F_1$ value over the DSIFN-CD dataset with other SOTA models. 
\par
GRAD-Former outperforms all other models in terms of all the evaluation metrics in
each dataset (Table \ref{tab:ALL}). On the CDD dataset, our model achieves an $F_1$ score of 97.57\%, an IoU of 95.26\%, and an OA of 99.43\%, surpassing all previous methods. Compared to the best Transformer-based model, ScratchFormer, our approach improves the $F_1$ score by 1.45\% and the IoU by 2.72\%, indicating superior feature extraction capabilities. When compared to the best CNN-based model, SNUNet, our method achieves a 2.23\% higher IoU and a 0.53\% gain in OA, highlighting its ability to capture important spatial and contextual relationships more effectively as depicted in table.

On the DSIFN-CD dataset, our method maintains its superiority, obtaining the highest $F_1$ score of 93.14\%, an IoU of 87.16\%, and an OA of 97.65\%. In comparison to ChangeMamba, the second-best model in this dataset, our model achieves a 2.93\% improvement in $F_1$ score and about 5.0\% increase in IoU. Moreover, it outperforms ChangeFormer, the leading Transformer-based model, by 1.89\% in OA. These results indicate that our model effectively handles complex changes and lower spatial resolution scenarios where CNN-based and SSM-based architectures struggle.

On the LEVIR-CD dataset, our approach continues to demonstrate outstanding performance with an $F_1$ score of 91.52\%, an IoU of 84.36\%, and an OA of 99.14\%. It surpasses CICD \cite{cicd}, the best-performing Transformer-based model, by 0.49\% in IoU and 0.10\% in OA. When compared to Mamba-based models, our method outperforms CDMamba by 1.27\% in IoU and 1.11\% in $F_1$ score, indicating that our modifications further enhance long-range dependency modeling and relevant feature aggregation.

Notably, GRAD-Former does not use any pre-trained backbone feature extractor and still outperforms the state-of-the-art method with fewer trainable parameters, demonstrating superior generalization across diverse datasets. These results highlight the robustness of our approach in capturing relevant spatial and contextual relationships across all different VHR datasets. 

\begin{figure*}[t]
\begin{center}
   \includegraphics[width=1\linewidth]{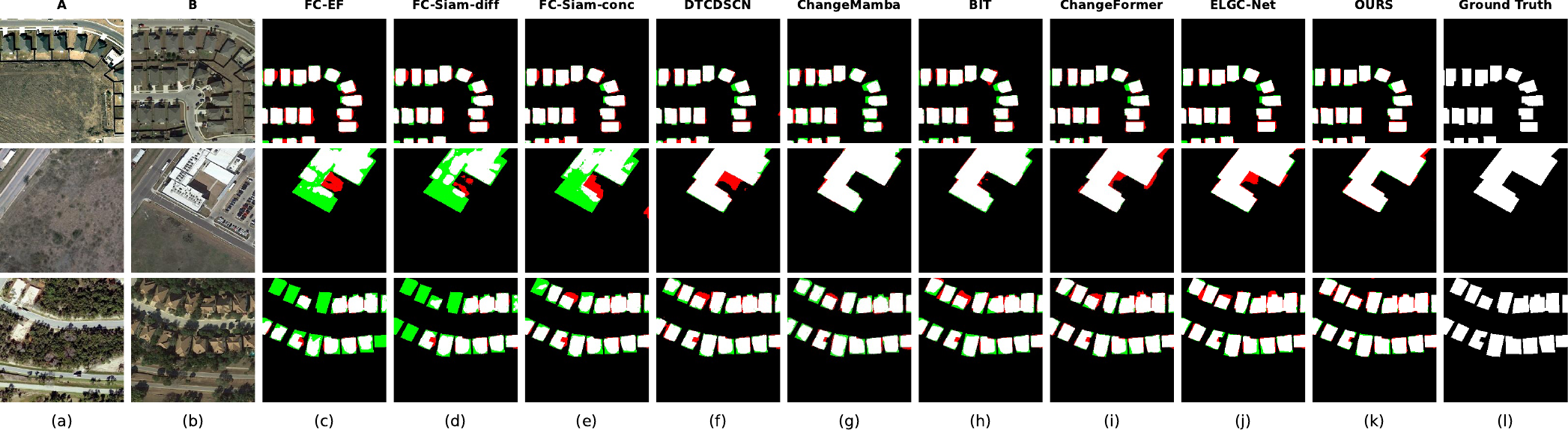}
\end{center}
   \caption{Qualitative comparisons on the LEVIR-CD dataset. We show the comparison with the eight best existing change detection
approaches in the literature, whose codebases are publicly available.(a) Pre-Change image (A), (b) Post-Change image (B), (c) FC-EF \cite{18}, (d) FC-Siam-diff \cite{18}, (e) FC-Siam-conc \cite{18}, (f) DTCDSCN \cite{DTCDSCN}, (g) ChangeMamba \cite{changemamba}, (h) BIT \cite{bit}, (i) ChangeFormer \cite{bit}, (j) ELGC-Net \cite{elgcnet}, (k) GRAD-Former (OURS) and (l) Ground truth. Color scheme: white
represents TP (i.e., “changed”), black corresponds to TN (i.e., “unchanged”), \textcolor{darkred}{red} indicates FP, and \textcolor{darkgreen}{green} denotes FN. }
\label{fig:levir}
\end{figure*}

\subsection{Qualitative Comparison with State-of-The-Art Methods}
\par
\begin{figure*}[t]
\begin{center}
   \includegraphics[width=1\linewidth]{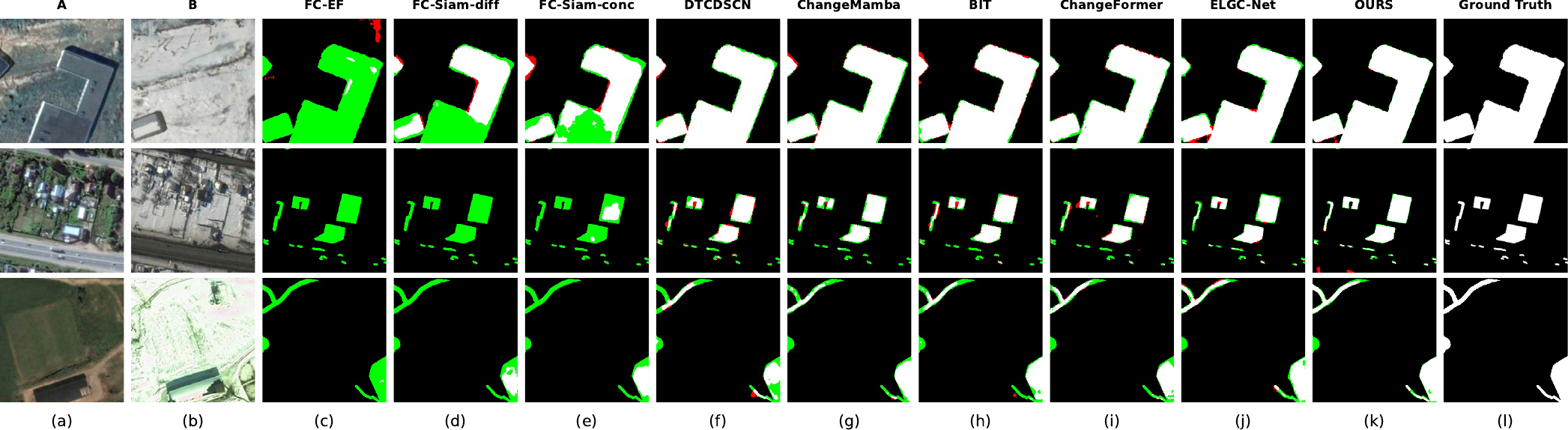}
\end{center}
   \caption{Qualitative comparisons on the CDD dataset. We show the comparison with the eight best existing change detection
approaches in the literature, whose codebases are publicly available.(a) Pre-Change image (A), (b) Post-Change image (B), (c) FC-EF \cite{18}, (d) FC-Siam-diff \cite{18}, (e) FC-Siam-conc \cite{18}, (f) DTCDSCN \cite{DTCDSCN}, (g) ChangeMamba \cite{changemamba}, (h) BIT \cite{bit}, (i) ChangeFormer \cite{bit}, (j) ELGC-Net \cite{elgcnet}, (k) GRAD-Former (OURS) and (l) Ground truth. Color scheme: white
represents TP (i.e., “changed”), black corresponds to TN (i.e., “unchanged”), \textcolor{darkred}{red} indicates FP, and \textcolor{darkgreen}{green} denotes FN. }
\label{fig:cdd}
\end{figure*}

 Fig. \ref{fig:dsifn}, Fig. \ref{fig:levir} and Fig. \ref{fig:cdd} present a comparative visualization of various SOTA methods on DSIFN-CD, LEVIR-CD and CDD datasets respectively as \cite{mf-vmamba}. The results indicate that U-Net-based approaches exhibit a higher prevalence of false negatives and false positives. However, integrating attention mechanisms significantly mitigates these issues. While these methods can generally identify all changed areas, they still struggle with detecting small-area samples and face challenges in accurately recognizing changes. This difficulty may arise due to irrelevant variations introduced by seasonal differences and lighting conditions. Additionally, since the sample size of images containing these objects, such as vehicles, is relatively small, the detected change regions often remain incomplete and inaccurate.
\par GRAD-Former solves this issue of pseudo-changes while providing clearer boundaries for detected objects and improving inter-object semantic consistency. The results from all three datasets highlight that GRAD-Former achieves superior visual outcomes compared to other models. The proposed method effectively reduces false negatives by leveraging gated mechanisms to identify the important features and eliminate noise along with global context modeling through attention. Observing Fig. \ref{fig:dsifn} our model produces binary change maps nearly indistinguishable from the ground truth and is able to detect sharp edges and distinguish between areas of small and large change. In Fig. \ref{fig:levir} one can observe that our model minimizes the false positives and detects minute details between the changed areas in scene 3 Fig. \ref{fig:levir}. As in Fig. \ref{fig:cdd} our model is able to ignore all the seasonal changes and come up with highly accurate change maps (in scene 1 Fig. \ref{fig:cdd}) as compared to other SOTA models, our model is the only model that is able to identity the stream on the left (in scene 2 Fig. \ref{fig:cdd}) accurately and a lot of the smaller changed areas attributing to our model's robustness and its ability to focus on relevant changed areas and ignore non-important features and noise.

\subsection{Ablation Study}

\subsubsection{Ablation of feature extraction modules in AFRAR module and fusion module}
We assess the impact of our proposed methods, including the SEA, the GLFR, and the DA module, on the LEVIR-CD dataset. To achieve this, we systematically apply various combinations of these techniques and report their corresponding parameters and IoU, $F_1$ scores and OA(\%) as shown in Table. \ref{tab:ablation_study}. Our final network, which integrates all contextual aggregation methods, achieves the highest IoU score of 84.36\%, with higher parameter efficiency, as presented in the last row of Table. \ref{tab:ablation_study}. This demonstrates that our module effectively cancels out the noise present in the VHR datasets, incorporating multiple contextual aggregation strategies in parallel.


\begin{table}[H]
\centering
\renewcommand{\arraystretch}{1.2}
\caption{Comparison of different methods by model size, computational requirements, and performance metrics. Our proposed method achieves the highest $F_1$ score (93.14\%) with a moderate parameter count.}
\label{tab:method_comparison}
\begin{tabular}{cccc}
\toprule
\textbf{Method} & \textbf{Params(M)} & \textbf{GFLOPs} & \textbf{$F_1$} \\
\midrule
FC-EF \cite{18} & 1.35 & 14.31 & 64.38 \\
FC-EF-conc \cite{18}& 1.54 & 21.32 & 62.03 \\
FC-EF-diff \cite{18} & 1.35 & 18.91 & 65.42 \\
DTCDSCN \cite{DTCDSCN}& 31.25 & 52.89 & 87.21 \\
SNUNet \cite{snu} & 12.03 & 21.33 & 79.64 \\
ISNet \cite{isnet} & 14.36 & 23.58 & 78.42 \\
SARAS-Net \cite{saras}& 56.89 & 139.90 & 67.58 \\
ChangeMamba \cite{changemamba} & 85.53 & 179.32 & 90.21 \\
RS-Mamba \cite{rs} & 52.73 & 95.74 & 86.67 \\
CDMamba \cite{cdmam} & 12.71 & 151.23 & 85.81 \\
MF-Vmamba \cite{mf-vmamba} & 57.84 & 94.06 & 88.05 \\
BIT \cite{bit}& 3.49 & 42.53 & 71.49 \\
ChangeFormer \cite{chnageformer}& 41.03 & 211.15 & 87.34 \\
ScratchFormer \cite{scratchformer} & 36.92 & 196.59 & 67.62 \\
ELGC-Net \cite{elgcnet} & 10.57 & 123.59 & 73.76 \\
\midrule
\textbf{GRAD-Former} & \textbf{10.90} & \textbf{129.50} & \textbf{93.14} \\
\bottomrule
\end{tabular}
\end{table}

\begin{table}[t]
\centering
\caption{Ablation study of combinations of Selective Embedding Amplification (SEA) module, Global-Local Feature Refinement (GLFR) module and the Difference Amalgamation (DA) module and their performance metrics.}
\label{tab:ablation_study}
\begin{tabular}{ccccccc}
\toprule
\multicolumn{3}{c}{\textbf{Module}} & \multirow{2}{*}{\textbf{Params (M)}} & \multirow{2}{*}{\textbf{IoU}} & \multirow{2}{*}{\textbf{$F_1$}} & \multirow{2}{*}{\textbf{OA(\%)}} \\
\cmidrule(lr){1-3}
\textbf{SEA} & \textbf{GLFR} & \textbf{DA} & & & & \\
\midrule
$\checkmark$ & $\times$ & $\times$ & 9.99 & 79.69 & 87.93 & 96.98 \\
$\times$ & $\times$ & $\checkmark$ & 10.72 & 80.30 & 88.9 & 97.09 \\ 
$\checkmark$ & $\times$ & $\checkmark$ & 10.71  & 81.22 & 88.99  & 97.80 \\ 
$\times$ & $\checkmark$ & $\times$ & 10.19 & 82.92 & 90.63 & 97.60 \\
$\times$ & $\checkmark$ & $\checkmark$ & 10.20 & 83.10 & 90.79 & 99.08 \\
$\checkmark$ & $\checkmark$ & $\times$ & 10.24 & 83.92 & 90.94 & 99.04 \\
$\checkmark$ & $\checkmark$ & $\checkmark$ & \textbf{10.90} & \textbf{84.36} & \textbf{91.52} & \textbf{99.14} \\
\bottomrule
\end{tabular}

\end{table}

\subsubsection{Ablation of different attention mechanisms in  GLFR module}
Our research confirms the efficiency of our differential attention mechanism, (as illustrated in Table \ref{tab:attention_comparison}). When we substituted standard self-attention for our differential attention in comparative testing, we found that differential attention not only requires fewer parameters but also improves the evaluation score. Our differential attention based module even outperformed  PT attention based module in terms of evaluation matrices and efficiency that claims to scale linearly. 
\begin{table}[htbp]
\centering
\caption{Comparison of different attention mechanisms in the GLFR module on LEVIR-CD dataset.}
\label{tab:attention_comparison}
\begin{tabular}{c|c|c|c|c}
\toprule
    
\textbf{Attention Type} & \textbf{Params (M)} & \textbf{$F_1$} & \textbf{IoU} & \textbf{OA(\%)} \\
\midrule
Simple Attention & 10.98 & 84.32 & 72.89 & 98.18 \\
Pooled-Transpose Attention & 10.83 & 91.08 & 83.62 & 99.1 \\
\textbf{Differential Attention} & \textbf{10.90} & \textbf{91.52} & \textbf{84.36} & \textbf{99.14} \\
\bottomrule
\end{tabular}
\end{table}

\subsubsection{Comparison of different losses  applied during training of GRAD-Former }
We also conducted an ablation study to examine how various loss functions affect the training of our proposed model. The results in Table \ref{tab:loss} clearly indicate that cross-entropy (CE) loss provides superior performance for our model training compared to alternative loss functions. As also presented by Noman \textit{et al.} \cite{elgcnet}, cross-entropy's stable gradient dynamics and pixel-wise optimization align effectively with siamese architecture's spatial feature extraction. Conversely, focal loss over-emphasizes hard negatives, causing training instability, while mIoU loss suffers from gradient instability when change regions are sparse, resulting in suboptimal convergence for change detection tasks.

\begin{table}[h!]
\centering
\caption{Comparative analysis with different loss functions applied during the model training on LEVIR-CD dataset, with optimal performance results emphasized in bold.}
\label{tab:loss}
\begin{tabular}{c|c|c|c}
\hline
\textbf{Loss Function} & \textbf{$F_1$} & \textbf{IoU} & \textbf{OA(\%)} \\
\hline
Focal Loss & 84.32 & 72.89 & 98.18 \\
mIoU Loss & 48.93 & 47.90 & 95.80 \\
\textbf{Cross-Entropy Loss} & \textbf{91.52} & \textbf{84.36} & \textbf{99.14} \\
\hline
\end{tabular}
\end{table}
\section{Conclusion}
\label{sec:V}
This paper presents GRAD-Former, a model designed for remote sensing change detection, which addresses the under-utilization of relevant information in very high-resolution (VHR) satellite images through its novel AFRAR module. GRAD-Former efficiently integrates global-local contextual feature aggregators and gating modules to enhance semantic information, while maintaining a minimal number of trainable parameters. Notably, it operates without the need for a pre-trained backbone feature extractor. Extensive experiments validate the effectiveness of our approach, with GRAD-Former achieving state-of-the-art performance across three challenging datasets and setting new benchmarks. Future work will aim to further improve efficiency and adapt the model for real-time change detection on edge devices.
\bibliographystyle{IEEEtran}
\bibliography{egbib}

\end{document}


\title{\Large\textbf{GRAD-Former: Gated Robust Attention-based Differential Transformer for Change Detection}\\[1em]
\large {Supplementary Material}}
\footnotetext[1]{These authors contributed equally to this work.}

\date{}
\maketitle

\thispagestyle{fancy}

\section{Introduction}
This document provides additional supporting results and analysis for our proposed GRAD-Former model for Change Detection (CD). The GRAD-Former framework enhances contextual understanding while maintaining efficiency through reduced model size, consisting of a novel encoder with Adaptive Feature Relevance and Refinement (AFRAR) module, fusion and decoder modules. This supplementary material presents a comprehensive quantitative and qualitative evaluation of our model's performance compared to state-of-the-art approaches, with particular focus on the WHU-CD dataset.

\section{Dataset Description}
The WHU-CD dataset \cite{whu} is an open-source collection focused on building change detection, was developed by researchers at Wuhan University using aerial imagery from Christchurch, New Zealand. It contains pre- and post-change images captured between 2012 and 2016, with a spatial resolution of 0.3 m/pixel. Covering an area of 20.5 km², the dataset documents urban redevelopment, featuring 12,796 buildings in 2012 and 16,077 in 2016, reflecting changes following the earthquake. For experimental purposes, the images are cropped to 256 × 256 pixels, with 5,947, 744, and 743 image pairs allocated for training, validation, and testing, respectively.

\section{Quantitative Performance Analysis}
We present a comprehensive comparative evaluation of GRAD-Former against existing state-of-the-art methodologies for remote sensing change detection on the WHU-CD dataset in Table S \ref{tab:comparison}. Our experimental results demonstrate that GRAD-Former consistently outperforms all other models across all evaluation metrics ($F_1$: 91.29\%, IoU: 83.98\%, OA: 99.31\%). With only 10.90M parameters, our model demonstrates significant computational efficiency compared to alternative high-performing architectures like BIT (57.84M), RS-Mamba (52.73M), and ChangeMamba (85.53M).

Compared to the second-best performing model RS-Mamba, our approach improves the $F_1$ score by 0.26\% and the IoU by 0.55\%, indicating superior feature extraction capabilities. When compared to BIT, which achieves the second-highest IoU value, GRAD-Former demonstrates a 0.54\% improvement, highlighting its ability to capture important spatial and contextual relationships more effectively. These results underscore the efficacy of our proposed Adaptive Feature Relevance and Refinement (AFRAR) module in enhancing the model's discriminative capacity while maintaining computational efficiency.

\begin{table}[h]
\centering
\begin{tabular}{|c|c|c|c|c|c|c|}
\hline
\multirow{2}{*}{\textbf{Type}} & \multirow{2}{*}{\textbf{Method}} & \multirow{2}{*}{\textbf{Params(M)}} & \multirow{2}{*}{\textbf{Publication}} & \multicolumn{3}{c|}{\textbf{WHU-CD}} \\
\cline{5-7}
 &  &  &  & \textbf{$F_1$} & \textbf{IoU} & \textbf{OA(\%)} \\
\hline
\multirow{7}{*}{\centering CNN-based} & FC-EF \cite{18}& 1.35 & \textit{ICIP-18} & 80.64 & 67.56 & 98.12 \\
 & FC-Sima-diff \cite{18}& 1.54 & \textit{ICIP-18} & 67.64 & 51.1 & 95.82 \\
 & FC-Sima-conc \cite{18}& 1.35 & \textit{ICIP-18} & 64.66 & 47.78 & 95.32 \\
 & DTCDSCN \cite{DTCDSCN} & 31.25 & \textit{LGRS-21} & 87.79 & 78.23 & 98.89 \\
 & SNUNet \cite{snu} & 12.03 & \textit{LGRS-22} & 90.29 & 82.3 & 99.05 \\
 & ISNet \cite{isnet}& 14.36 & \textit{TGRS-22} & 83.24 & 72.47 & 98.38 \\
 & SARAS-Net \cite{saras}& 56.89 & \textit{LGRS-23} & 87.09 & 77.14 & 98.89 \\
\hline
\multirow{2}{*}{\centering Mamba-based} & ChangeMamba \cite{changemamba} & 85.53 & \textit{TGRS-24} & 84.11 & 78.05 & 98.86 \\
 & RS-Mamba \cite{rs} & 52.73 & \textit{LGRS-24} & {\color{red}91.03} & {83.43} & {99.22} \\
\hline
\multirow{8}{*}{\centering Transformer-based} & BIT \cite{bit} & 57.84 & \textit{TGRS-22} & 90.97 & {\color{red}83.44} & 99.11 \\
 & ChangeFormer \cite{chnageformer} & 41.03 & \textit{IGARSS-22} & 85.31 & 79.02 & 99.00 \\
 & ScratchFormer \cite{scratchformer} & 36.92 & \textit{TGRS-24} & 87.78 & 78.22 & 98.93 \\
 & ELGC-Net \cite{elgcnet} & 10.57 & \textit{TGRS-24} & 87.54 & 77.85 & 98.84 \\
 & MSFNet \cite{msf} & N/A & \textit{JSTARS-24} & 90.87 & 82.8 & 99.07 \\
 & RaHFF-Net \cite{rahff} & 16.48 & \textit{JSTARS-25} & 90.43 & 81.72 & {\color{red}99.23} \\
 & AMDANet \cite{amda} & 25.83 & \textit{TGRS-25} & 89.3 & 80.68 & 99.03 \\
 & AdvCP \cite{adver} & N/A & \textit{TIP-26} & 77.25 & 62.39 & 98.51 \\
 & \textbf{GRAD-Former} & 10.90 & - & {\color{green}91.29} & {\color{green}83.98} & {\color{green}99.31} \\
\hline
\end{tabular}
\caption{Comparison of different methods for remote sensing change detection on the WHU dataset over $F_1$ score, IoU and Overall Accuracy(OA (\%)) where $F_1$ and IoU are calculated over the change class. The \textcolor{green}{highest} and \textcolor{red}{second highest} values in each metric column are highlighted.}
\label{tab:comparison}
\end{table}

\section{Computational Efficiency-Performance Trade-off Analysis}

\begin{figure}[h!]
\begin{center}
   \includegraphics[width=0.75\linewidth]{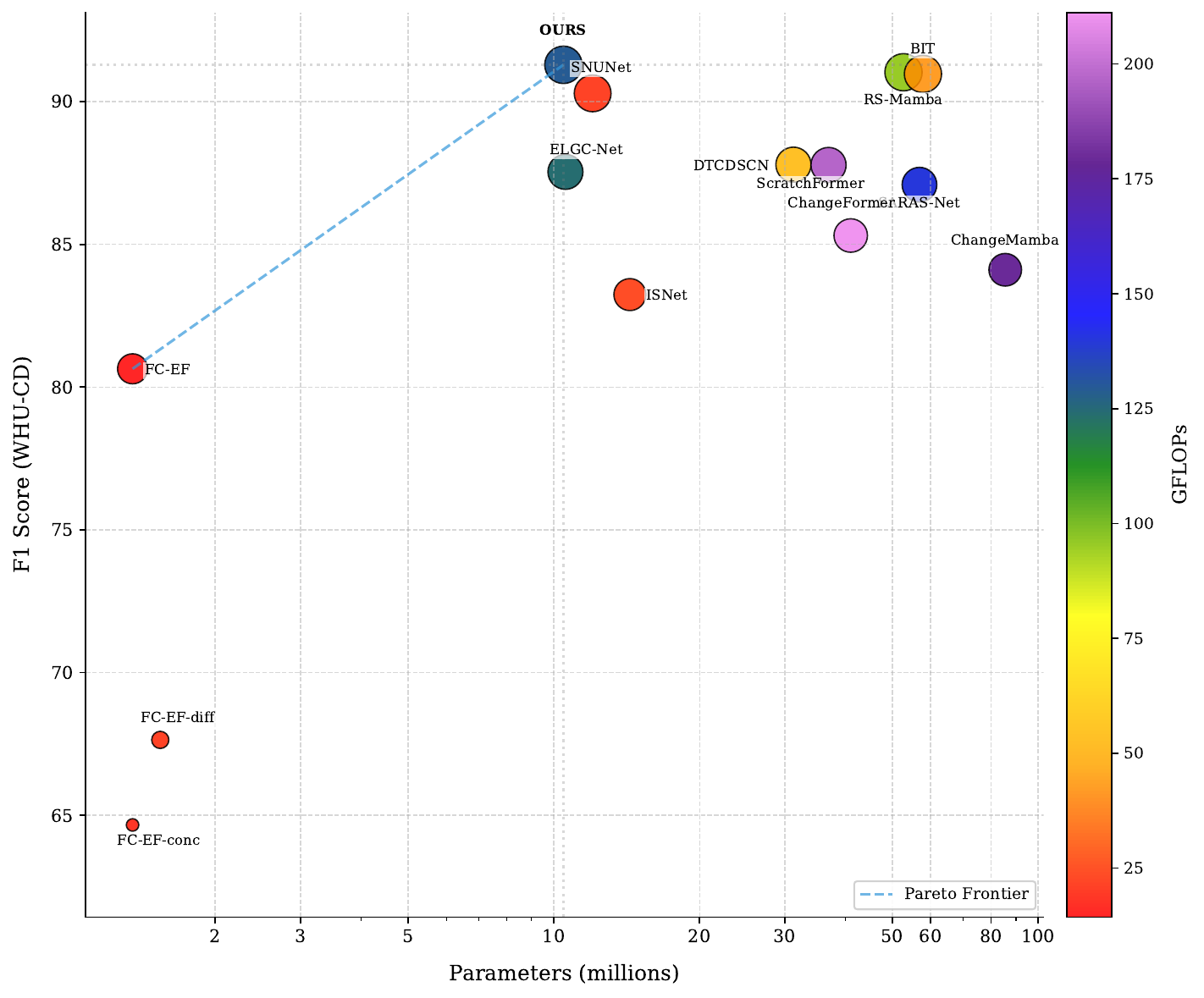}
\end{center}
   \caption{Performance comparison of change detection models as per Table \ref{tab:comparison} . The plot illustrates the trade-off between model size (parameters in millions) and performance ($F_1$ score over WHU-CD dataset). Bubble color indicates computational complexity (GFLOPs). The dashed blue line represents the Pareto frontier of optimal models. Lying on the Pareto's front our model achieves state-of-the-art $F_1$ score (91.29\%),while maintaining reasonable computational efficiency (129.5 GFLOPs).}
\label{fig:plot}
\end{figure}
Figure S \ref{fig:plot} provides a visualization of the efficiency-performance trade-off among comparative change detection models through a Pareto efficiency plot. GRAD-Former occupies a position on the Pareto frontier, indicating an optimal balance between architectural complexity and performance. Despite its superior discriminative capabilities, our model maintains acceptable computational demands (129.5 GFLOPs), facilitating practical deployment in resource-constrained environments.

The analytical visualization demonstrates that numerous competing methodologies either require substantially greater parameterization to achieve comparable performance or exhibit inferior metrics despite equivalent or greater model complexity. ChangeMamba \cite{changemamba} requires approximately 8 times more parameters than GRAD-Former while achieving lower $F_1$ scores, while BIT \cite{bit} necessitates about 5.5 times more parameters to achieve comparable but still inferior performance. This substantial parameter efficiency becomes particularly valuable in operational remote sensing contexts where computational resources may be limited. The relationship between computational complexity and performance further validates our approach, substantiating GRAD-Former's position as a computationally efficient yet high-performing architecture for remote sensing change detection applications.

\section{Qualitative Results Comparison}

\begin{figure}[h]
\begin{center}
   \includegraphics[width=\linewidth]{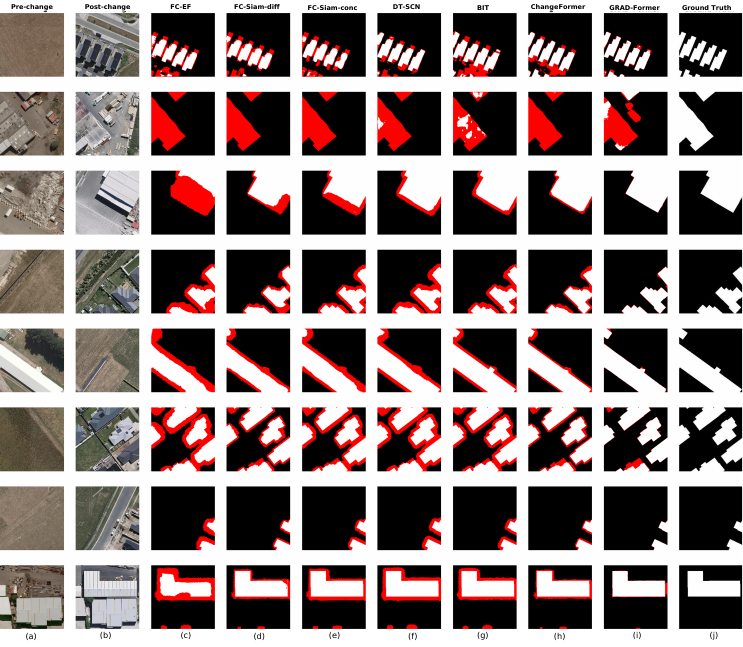}
\end{center}
   \caption{A visual comparison of various state-of-the-art change detection methods applied to the WHU-CD dataset: (a) pre-change image, (b) post-change image, (c) FC-EF \cite{18}, (d) FC-Siam-diff \cite{18}, (e) FC-Siam-conc \cite{18}, (f) DT-SCN \cite{DTCDSCN}, (g) BIT \cite{bit}, (h) ChangeFormer \cite{chnageformer}, (i) GRAD-Former, and (j) the ground truth. In the results, white pixels represent true positives (areas of change), black pixels represent true negatives (unchanged areas), and red pixels denote both false positives and false negatives.}
\label{fig:whu-comp}
\end{figure}

Figure S \ref{fig:whu-comp} presents a qualitative comparison of GRAD-Former against several state-of-the-art change detection methodologies using exemplar images from the WHU-CD dataset. GRAD-Former exhibits superior capability in precisely demarcating the geometric boundaries of changed regions compared to predecessor methodologies (FC-EF \cite{18}, FC-Siam-diff\cite{18}, FC-Siam-conc \cite{18}), which show significant boundary localization errors. Our proposed architecture demonstrates notably reduced occurrence of false positives and false negatives (represented by red pixels) compared to alternative approaches, indicating enhanced discriminative reliability.

GRAD-Former effectively captures complex structural changes that other methods often overlook, especially in areas with subtle semantic variations. Our qualitative analysis shows that the model delivers consistent detection performance across spatially diverse regions. In contrast, other transformer-based approaches show variable performance depending on the complexity of the semantic change. Earlier CNN-based methods struggle to capture fine-grained changes, while newer transformer-based models offer improvements but still tend to produce imprecise boundaries. GRAD-Former generates results that align closely with the ground truth, with well-defined building edges and minimal false detections. This highlights its ability to integrate both local and global contextual information effectively through the proposed AFRAR module.

\section{Conclusion}
The supplementary analysis presented in this document provides comprehensive empirical evidence of GRAD-Former's superior performance in remote sensing change detection tasks. Through rigorous quantitative comparative evaluation, computational efficiency analysis, and qualitative assessment, we demonstrate that our proposed architectural framework establishes a new benchmark for the domain. The integration of the novel Adaptive Feature Relevance and Refinement (AFRAR) module, with its constituent Selective Embedding Amplification (SEA) and Global-Local Feature Refinement (GLFR) components, enables GRAD-Former to achieve state-of-the-art performance while maintaining parameter efficiency.

These empirical findings validate our methodological approach of enhancing contextual understanding through global-local feature integration, and demonstrate that GRAD-Former effectively addresses the methodological challenges of precisely delineating change regions in very high-resolution satellite imagery.

\newpage
\clearpage
\bibliographystyle{IEEEtran}
\bibliography{reference}